\documentclass[11pt]{article}
\usepackage{amsmath}
\usepackage{amsfonts}
\usepackage{varwidth}
\usepackage{epsf}
\usepackage{algorithm}
\usepackage{algorithmic}
\usepackage{xspace}
\usepackage{booktabs}
\usepackage{eucal, bm}

\newcommand{\La}{\lambda}
\newcommand{\Ga}{\gamma}
\newcommand{\Ss}{\mathcal{S}}
\newcommand{\Aa}{\mathcal{A}}
\newcommand{\ra}{\rightarrow}
\newcommand{\la}{\leftarrow}
\newcommand{\RR}{\mathbb{R}}
\newcommand{\w}{{\bm w}}
\newcommand{\e}{{\bm e}}
\def\ph{\bm\phi}
\def\th{\bm\theta}

\title{GQ($\La$) \\
Quick Reference and Implementation Guide}
\author{Adam White and Richard S. Sutton}
\date{Revised July 29, 2014}

\begin{document}
\maketitle

This document should serve as a quick reference for and guide to the implementation of linear GQ($\La$), a gradient-based off-policy temporal-difference learning algorithm. Explanation of the intuition and theory behind the algorithm are provided elsewhere (e.g., Maei \& Sutton 2010, Maei 2011). If you questions or concerns about the content in this document or the attached java code please email Adam White (adam.white@ualberta.ca).

\section{Requirements and Setting}
For each use of GQ($\La$) you will need to provide three {\em question functions} specifying the quantity to be predicted, and four \emph{answer functions} characterizing the approximation that will be found. Let $\Ss$ and $\Aa$ denote the sets of states and actions. Then the question functions are:
\begin{itemize}
\itemsep-1em
\item $\pi: \Ss \times \Aa \ra [0,1]$; target policy to be learned. Incidently, if $\pi$ is chosen as the greedy policy with respect to the learned value function, then the algorithm will implement a generalization of the Greedy-GQ algorithm (Maei, Szepesvari, Bhatnagar \& Sutton 2010).\\
\item $\Ga : \Ss \ra [0, 1]$; termination or discounting function ($\Ga (s) = 1 - \beta(s)$ in GQ paper)\\
\item $r: \Ss \times \Aa \times \Ss \ra \RR$; reward function
\end{itemize}
In many publications there is also specified a fourth question function, the terminal reward function $z : \Ss \ra \RR$ used to specify a final reward at termination. More recently its has been recognized that this functionality can be included in the reward function, making use of the discounting function (Modayil, White \& Sutton 2014). For example, if one wanted only a terminal reward function $z(s)$ upon termination in state $s$, one would use a reward function of $r(s, a, s') = (1-\gamma(s'))z(s')$. This completes the specification of the predictive question that you are seeking to answer using the GQ($\La$) algorithm.

The answer functions are:
\begin{itemize}
\itemsep-1em
\item $b: \Ss \times \Aa \ra [0,1]$; behavior policy\\
\item $I : \Ss \times \Aa \ra [0, 1]$; interest function (can set to 1 for all state-action pairs or indicate selected state-action pairs to be best approximated)\\
\item $\ph : \Ss \times \Aa \ra \RR^n$; feature-vector function\\
\item $\La : \Ss \ra [0, 1]$; bootstrapping or eligibility-trace decay-rate function
\end{itemize}

The following data structures are internal to GQ:
\begin{itemize}
\itemsep-1em
\item $\th \in \RR^n$; the learned weights of the linear approximation: $Q^\pi(s,a) = \th^\top\ph(s, a) = \sum_{i=1}^n \th_i\ph_i(s,a)$\\
\item $\w \in \RR^n$; secondary set of learned weights\\
\item $\e \in \RR^n$; eligibility trace vector
\end{itemize}
Parameters internal to GQ:
\begin{itemize}
\item $\alpha$; step-size parameter for learning $\th$
\item $\eta \in [0, 1]$; relative step-size parameter for learning $\w$ $(\alpha\eta)$
\end{itemize}

\section{Algorithm Specification}

We can now specify GQ($\La$). Let $\w$ and $\e$ be initialized to zero and $\th$ be initialized arbitrarily. Let the subscript $t$ denote the current time step. Let $\rho_t$ denote the ``importance sampling" ratio:
\begin{equation}
\rho_t = \frac{\pi(S_t,A_t)}{b(S_t, A_t)}, 
\end{equation}
where $S_t$ and $A_t$ are the state and action occuring on time step $t$.
Let $\bar\ph_t$ denote the expected next feature vector, defined by:
\begin{equation}
\bar\ph_t = \sum_{a \in \Aa} \pi(S_t,a)\ph(S_t,a) 
\end{equation}
\goodbreak\noindent
Then the following equations fully specify GQ($\La$):
\begin{equation}
\delta_t = r(S_t,A_t,S_{t+1}) + \Ga(S_{t+1})\th^\top_t \bar\ph_{t+1} - \th^\top_t \ph(S_t,A_t) 
\end{equation}
\begin{equation}
\th_{t+1} = \th_{t} + \alpha\left[\delta_t\e_t - \Ga(S_{t+1})(1 - \La(S_{t+1}))(\w^\top_t \e_t)\bar\ph_{t+1}\right] 
\end{equation}
\begin{equation}
\w_{t+1} = \w_t + \alpha\eta[\delta_t\e_t  - (\w^\top_t \ph(S_t,A_t))\ph(S_t,A_t)]
\end{equation}
\begin{equation}
\e_t = I(S_t)\ph(S_t,A_t) + \Ga(S_t) \La(S_t)\rho_t\e_{t-1} 
\end{equation}

\section{Pseudocode}

The following pseudocode characterizes the algorithm and its use.
\bigskip

\noindent\fbox{
\begin{varwidth}{\dimexpr\linewidth-2\fboxsep-2\fboxrule\relax}
\begin{tabbing}
Initialize $\th$ arbitrarily and $\w = 0$ \\
Repeat \=(for each episode):\\
\> Initialize $\e = 0$\\
\> $S \la$ initial state of episode \\
\> Repeat \=(for each step of episode):\\
\>\>$A \la$ action selected by policy $b$ in state $S$ \\
\>\>Take action $A$, observe next state, $S^\prime$\\
\>\> $\bar\ph \la 0$\\
\>\> For \=all $a \in \Aa(s)$:\\
\>\>\> $\bar\ph \la \bar\ph + \pi(S^\prime,a)\ph(S^\prime,a)$\\ 
\>\>$\rho = \frac{\pi(S, A)}{b(S, A)}$\\
\>\>GQlearn($\ph(S,A), \bar\ph, \La(S^\prime), \Ga(S'), r(S, A, S^\prime), \rho, I(S)$)\\
\>\>$S \la S^\prime$\\
\> until $S^\prime$ is terminal
\end{tabbing}
\end{varwidth}
}

\bigskip
\noindent\fbox{
\begin{varwidth}{\dimexpr\linewidth-2\fboxsep-2\fboxrule\relax}
\begin{tabbing}
GQ\=Learn($\ph,\bar\ph,\La,\Ga,R,\rho,I$)\\
\> $\delta \la R + \Ga\th^\top \bar\ph - \th^\top \ph$  \\
\> $\e \la \rho\e + I\ph$ \\
\> $\th \la \th + \alpha(\delta\e - \Ga(1 - \La)(\w^\top\e)\bar\ph)$\\
\>$\w \la \w + \alpha\eta(\delta\e - (\w^\top \ph)\ph)$\\
\>$\e \la \Ga\La\e$ 
\end{tabbing}
\end{varwidth}
}

\section{Code}
The files {\tt GQlambda.java} and {\tt GQlambda.cpp} (in the arXiv source archive) contain implementations of the GQlearn function described in the pseudocode. We have excluded optimizations (e.g., binary features or efficient trace implementation) to ensure the code is simple and easy to understand. We leave it to the reader to provide environment code for interfacing to GQ($\La$) (e.g., using RL-Glue).

\section{References}
\parindent=0pt
\def\hangin{\hangindent=0.15in}
\parskip=6pt

\hangin
Maei, H. R., Szepesv\'ari, Cs., Bhatnagar, S., Sutton, R. S. (2010). Toward off-policy learning control with function approximation. In {\em Proceedings of the 27th International Conference on Machine Learning}, Haifa, Israel.

\hangin
Maei, H. R. and Sutton, R. S. (2010). GQ($\La$): A general gradient algorithm for temporal-difference prediction learning with eligibility traces. In {\em Proceedings of the Third Conference on Artificial General Intelligence}, pp.~91--96.

\hangin
Modayil, J., White, A., Sutton, R.~S. (2014). Multi-timescale nexting in a reinforcement learning robot. \emph{Adaptive Behavior 22}(2):146--160.

\hangin
Sutton, R. S., Barto, A. G. (1998). Reinforcement Learning: An Introduction. MIT Press.
\end{document}